\documentclass[letterpaper, 10 pt, conference]{ieeeconf}  

\usepackage{algorithmic}
\usepackage{epsfig}
\usepackage{graphicx}
\usepackage{amsmath}
\usepackage{amssymb}
\usepackage{booktabs}
\usepackage{multirow}
\usepackage{multicol}
\usepackage[ruled,vlined]{algorithm2e}
\usepackage{xcolor}
\usepackage{csquotes}
\usepackage{pifont}
\usepackage{epigraph} 
\IEEEoverridecommandlockouts                              

\usepackage{booktabs}                                   
\usepackage{multirow}                                   
\usepackage{makecell}                                   
\usepackage{tablefootnote}                              
\usepackage[symbol]{footmisc}                           
\usepackage{amsmath,amssymb}                            
\usepackage{xcolor}                                     
\let\labelindent\relax              
\usepackage{enumitem}                                   
\usepackage{subcaption}                                 
\usepackage{stfloats}                                   
\usepackage[misc]{ifsym}     
\usepackage{hyperref}   

\newcommand{\eat}[1]{}                                  

\title{\LARGE \bf
Object-Centric Instruction Augmentation for Robotic Manipulation
}

\author{Junjie Wen$^{1,*}$, Yichen Zhu$^{2,*}$, Minjie Zhu$^{1}$, Jinming Li$^{3}$, Zhiyuan Xu$^{2}$, Zhengping Che$^{2}$,\\
Chaomin Shen$^{1,\dagger}$, Yaxin Peng$^{3}$, Dong Liu$^{2}$, Feifei Feng$^{2}$, and Jian Tang$^{2,\dagger}$
\thanks{$^{1}$School of Computer Science, East China Normal University, China
        {\tt\small \{51255901019, 51255901028\}@stu.ecnu.edu.cn, cmshen@cs.ecnu.edu.cn}}
\thanks{$^{2}$Midea Group, China
        {\tt\small \{zhuyc25, xuzy70, chezp, liudong13, feifei.feng, tangjian22\}@midea.com}}
\thanks{$^{3}$Department of Mathematics, School of Science, Shanghai University, China
        {\tt\small \{ljm2022, yaxin.peng\}@shu.edu.cn}}
\thanks{
    $^*$Equal contributions. This work was done during Junjie Wen, Minjie Zhu, and Jinming Li's internship at Midea Group.
}
\thanks{
    $^\dagger$Corresponding authors: Chaomin Shen and Jian Tang.
}
}

\begin{document}

\maketitle
\thispagestyle{empty}
\pagestyle{empty}

\begin{abstract}
Humans interpret scenes by recognizing both the identities and positions of objects in their observations. For a robot to perform tasks such as \enquote{pick and place}, understanding both what the objects are and where they are located is crucial. While the former has been extensively discussed in the literature that uses the large language model to enrich the text descriptions, the latter remains underexplored. In this work, we introduce the \textit{Object-Centric Instruction Augmentation (OCI)} framework to augment highly semantic and information-dense language instruction with position cues. We utilize a Multi-modal Large Language Model (MLLM) to weave knowledge of object locations into natural language instruction, thus aiding the policy network in mastering actions for versatile manipulation. Additionally, we present a feature reuse mechanism to integrate the vision-language features from off-the-shelf pre-trained MLLM into policy networks. Through a series of simulated and real-world robotic tasks, we demonstrate that robotic manipulator imitation policies trained with our enhanced instructions outperform those relying solely on traditional language instructions. The project is available at \href{https://oci-robotics.github.io/}{https://oci-robotics.github.io/}.
\end{abstract}

\section{Introduction}

A fundamental challenge in deciphering the intelligence of embodied agents lies in the representation and interpretation of the rich, continuous sensory data obtained from our environment. The \enquote{what-where} framework in cognitive science~\cite{ungerleider1994whatwhere, de2011usefulness} posits that our brain employs separate neural channels to encode two primary categories of information~\cite{hebart2012visual}. Specifically, the \enquote{what} pathway \cite{goodale1992separate} identifies detailed attributes of objects, like their identity and characteristics (e.g., color and shape). In contrast, the \enquote{where} pathway \cite{sheth2016two} discerns spatial details, such as position, trajectory, and closeness. Modern advancements in the domain of object-centric representations in artificial intelligence are probing these concepts \cite{freud2017large}, with a growing emphasis on simultaneously training both the \enquote{what} and \enquote{where} facets within specific contexts.

The recent advancement of the Large Language Model (LLM) has attracted increasing attention, with many initiatives leveraging human-like language comprehension for robotics. It revolutionized the pipeline for embodied agents and resolved the challenge in the previous literature where human instruction can be hard to interpret by robots~\cite{vemprala2023chatgpt, ahn2022saycan}. However, these methodologies primarily focus on augmenting the text with more detailed instructions, i.e., task planning~\cite {mu2023embodiedgpt, ahn2022saycan, xiao2022robotic}. The presumption is that by learning from the input observation, the policy networks should comprehend the objects' position internally and store this information as the form of parameters, and subsequently generate a corresponding action trajectory. While this is a prevalent approach, it often demands a vast number of demonstrations for the policy network to genuinely understand an object's position and produce a valid trajectory.

Our work is built upon the recent development of Multi-modal Large Language Models (MLLM) to augment language instruction with object-centric information. The novelty of our framework is how we inform the robot with visual information via transferred language instruction that allows us to infer high-level features around affordances and intents.


To start, we aim to develop a position-aware MLLM capable of representing the desired objects' locations. Our investigation delves into two kinds of positions: absolute and relative. The absolute position signifies an object's exact location. Taking the user instruction  
\enquote{pick up the apple and place it on the plate} as an example, we can directly extract the bounding boxes of both the apple and the plate, then supplement the instruction with these bounding boxes. This provides the policy network with a clear understanding of object locations based on a first-person perspective. Furthermore, any distractors, i.e., objects not pertinent to the action, are excluded, sharpening the policy network's focus on the task at hand. On the other hand, relative positions place the robot at the world's center, and define the direction of target objects relative to the robot. Using the previous example, our system might indicate that the \enquote{apple is to the left} and the \enquote{plate is to the right}. This mirrors the human-centric approach to perceiving object positions. These language augmentations can be perceived as visual cues presented in text, allowing the policy network to derive positional knowledge from natural language. This circumvents the challenges associated with learning about objects directly from images.

Moreover, considering the computational intensity of MLLM calculations, we have devised a method that repurposes MLLM features, marrying its vision-language insights with policy networks to enhance robotic manipulation performance. By caching the MLLM's pre-computed embedding at the onset of the inference phase, we enable policy networks to operate seamlessly and efficiently, leveraging MLLM knowledge without incurring high computational costs.

\begin{figure*}[t]
    \centering
    \includegraphics[width=0.80\textwidth]{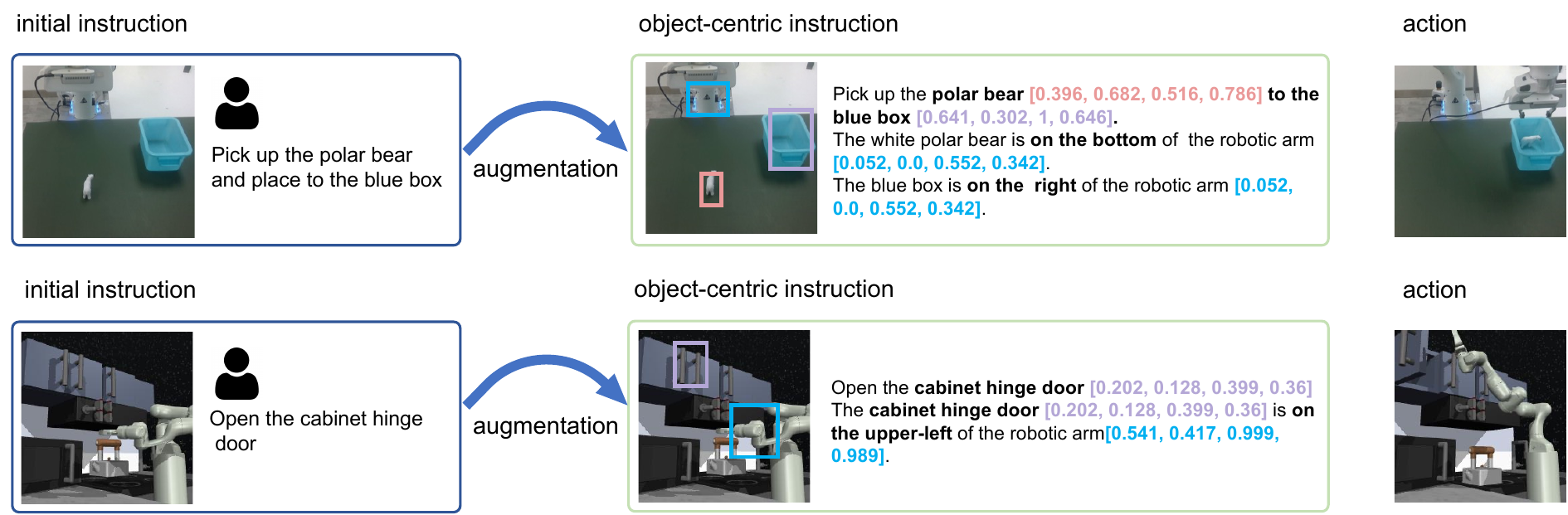}
    \caption{Two examples of object-centric instruction augmentation for simulation and real robots, respectively. Given an initial instruction from the left figure, we augment them by providing the object's absolute position and relative position to the robotics and obtain the action eventually.}
    \label{fig:example}
\end{figure*}

We assessed our proposed Object-Centric Instruction augmentation (OCI) on unseen simulated and real-world robotic manipulation settings. Our findings suggest that OCI not only facilitates superior policy learning in contrast to flat instructions, but also enhances the understanding of the pivotal influence of an object's position on the success rate of manipulation. Through detailed ablation studies, we emphasize the importance of using pretrained models in \enquote{where} dimensions as vital for achieving remarkable enhancements.

\noindent
\textbf{To summarize, our contributions are the follows:}
\begin{itemize}
    \item We've established a pipeline that equips MLLM with position-aware knowledge, enabling it to augment incoming instructions with object-centric information. 
    \item We introduce a Feature Reuse Mechanism that incorporates feature embedding from MLLM into the policy networks. 
    \item Empirical evaluations and ablations on both simulated and real robots validate the superiority of our framework over existing baselines. 
\end{itemize}
\section{Related Work}

\noindent
\textbf{Natural Language Instruction for Robotic Manipulation.} The rise of large foundation models, specifically the Large Language Model (LLM)~\cite{cot, clip, zs_reasoner, fan2022minedojo, shridhar2022cliport}, has opened a new era for embodied agents to understand instruction and generalize to unseen domains. These works include task planning~\cite{huang2022language, maskill}, navigation~\cite{visualmapnav}, code generation for actions~\cite{cop}, multi-robot collaboration~\cite{mandi2023roco}, human-robot interaction~\cite{cui2023no}, etc,. Another line of work utilize vision-language models for robotics~\cite{driess2023palm}. Some of them use pre-trained VLM as instruction encoder~\cite{jiang2022vima, hill2020human, jang2022bcz, brohan2022rt1, nair2022learning, lynch2020language, shridhar2023perceiver} or visual encoder~\cite{ma2022vip, laskin2020reinforcement, shah2021rrl, shridhar2022cliport}, and others for visual state representations~\cite{karamcheti2023language}, object navigation~\cite{gadre2022clip, moo}, high-level planning~\cite{driess2023palm, mu2023embodiedgpt, nlmap}, or providing supervision or success detection~\cite{du2023vision, xiao2022robotic, zhang2023grounding, sumers2023distilling}, and open-vocabulary object localization~\cite{moo}. RT-2~\cite{rt2} develop an end-to-end framework that outputs actions with images and instructions directly. 

Unlike prior applications of foundational models to downstream tasks like planning and navigation, our goal is to enhance language instructions to boost the generalizability of robotic manipulation, emphasizing positional cues for objects in text structure.

\noindent
\textbf{Object-Centric Representation Learning.} The fields of robotics and vision have delved deeply into the use of object-centric representations, recognizing their value in facilitating modular reasoning about visual scenes. Within the robotics domain, it is common to use poses~\cite{tremblay2018deep, tyree20226} and bounding boxes~\cite{migimatsu2020object, wang2019deep} to represent objects presented in a scene. However, these methods are typically limited to predefined object categories or specific instances. The development of the vision foundation model, i.e., SAM~\cite{kirillov2023segment}, motivates the researcher to extend the object-centric learning~\cite{zhu2023learning, shiplug} to unseen objects.  Some works leverage unsupervised learning~\cite{locatello2020object, burgess2019monet} approaches to endow the manipulation policies with object awareness~\cite{wang2021generalization, heravi2023visuomotor}, but their works are limited to simulation. Unlike prior works, we present a novel approach that translates object positions into natural language descriptions, enhancing visual representations for more effective policy learning in manipulation tasks. This method aligns with recent trends where LLM are integrated with vision systems~\cite{minigpt-4, llava, LLaVA-phi}, resulting in enriched scene comprehension through a multi-modal approach.

\begin{figure*}[t]
    \centering
    \includegraphics[width=0.85\textwidth]{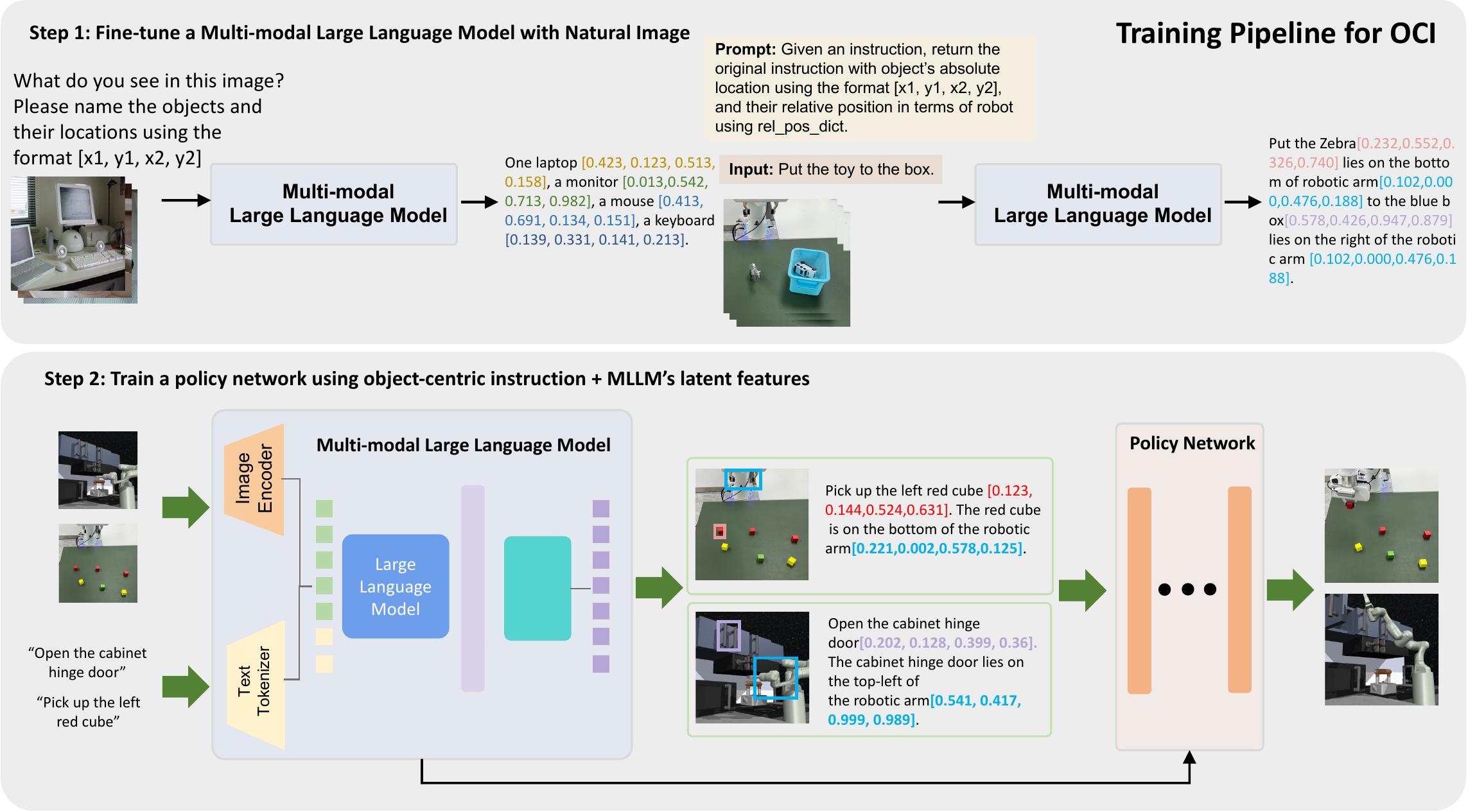}
    \caption{An overview of the training process in OCI. We first fine-tune a MLLM with general detection datasets and our collected datasets. The fine-tuned MLLM enables automatically augmenting instruction object-centric information. Subsequently, this is integrated with a policy network to develop a model capable of generating specific actions.}\label{fig:oci_train}
\end{figure*}

\section{Methodology}
This section discusses how to chain what and where into a unified and useful instruction for manipulation policy learning. There are two key components to our OCI framework: 1) a fine-tuned MLLM that is adept at comprehending language and the environment, with the ability to correlate an object's location to its identity, and 2) a feature reuse mechanism that leverages the features embedding from MLLM to improve policy learning. The following section will address these two challenges in detail. 

\subsection{Position-Aware Multi-modal Large Language Models}

\noindent
\textbf{Finetune Position-Aware MLLM.} We use the pretrained weights~\cite{minigpt-4} that were trained on a combined dataset of Conceptual Caption, SBU~\cite{sbu}, and LAION~\cite{schuhmann2022laion}. The entire pre-training stage undergoes 20,000 training steps with a batch size of 256, covering approximately 5 million image-text pairs. To enhance its visual comprehension of text queries, we fine-tuned the pre-trained models using the annotated LLaVA-Instruct-150K dataset~\cite{llava}. While the VLM currently demonstrates competent vision-language integration at the image level, this is inadequate for our purposes. We aim for scene comprehension that transcends mere image-level analysis, emphasizing object-centric representation from a holistic standpoint.

To facilitate the MLLM with object-level understanding and comprehend objects in terms of location, we use referring detection datasets such as RefCOCO, RefCOCO+, and RefCOCOg~\cite{yu2016modeling} to fine-tune the model. Notice that the RefCOCO datasets contain object's bounding boxes, accompanied by questions and answers. These datasets identify bounding boxes of mentioned objects within given instructions. This closely aligns with our use-case, which requires understanding the absolute position of objects.

Now, the MLLM can return the bounding boxes of referred objects but still cannot re-write the given instruction. We further make an instructing tuning dataset that allows the model to know how to rewrite the incoming command into our desired format, which preserves the instruction, adds bounding boxes to the target objects, and provides the relative position of the objects. In particular, we formulate the format of prompt, input, and output. We use a set of data that contains these three components, along with the image for the current scene, to fine-tune the MLLM. We collected 200 number of high-quality data for this fine-tuning step. Each image is collected with a random setup on the table. The initial input language is manually set, and we use the GPT-3.5-turbo~\cite{ouyang2022training} to enrich the instruction format. We use $o_{1}$ and $o_{2}$ to represent two objects for simplicity. In practice, we use a concrete object's name. We ask the GPT to \enquote{Use three different expressions to rewrite the instruction of 'pick up $o_{1}$ and place to $o_{2}$'.}, it returns \enquote{Grab $o_{1}$ and set it on $o_{2}$.}, \enquote{Take $o_{1}$ and position it on $o_{2}$.}, and \enquote{Lift $o_{1}$ and put it onto $o_{2}$}. In our experiment, each instruction is augmented with ten more different expressions, which are then used for the training. 

\noindent
\textbf{Representation of Position.} We represent the absolute position of objects using bounding boxes. These positions are articulated with numerical values, expressed in a manner that's both intuitive and aligned with natural language conventions. A bounding box is characterized by the format $[x_{\min}, y_{\min}, x_{\max}, y_{\max}]$. Here, both $x$ and $y$ are normalized in relation to the image's dimensions. By default, coordinates are presented with a precision of three decimal places. Such coordinates can appear at any point within the model's input or output sequence.

While bounding boxes define the absolute position of objects, a description of their relative positioning remains essential. To articulate the relative position of an object, a reference point is necessary. In the context of robotic manipulation, understanding movement in relation to the robot's own position is pivotal. As such, we designate the robot as the central reference in the image and detail objects' positions relative to it. For this purpose, we provide eight directional options: (a) Left, (b) Right, (c) Top, (d) Bottom, (e) Upper-left, (f) Upper-right, (g) Bottom-left, (h) Bottom-right. 

Overall, when providing linguistic instructions to the robot, our pipeline facilitates object-centric commands. These commands guide the policy network with specific positional expressions. For instance, as shown in Figure~\ref{fig:example}, when the user instructs the robot to \enquote{pick up the polar bear and place it in the blue box}, the following instructions with coordinates are given: \enquote{Pick up the polar bear [0.396, 0.682, 0.516, 0.786] to the blue box [0.641, 0.302, 1, 0.646]. The white polar bear is on the bottom of the robotic arm [0.052, 0.0, 0.552, 0.342]. The blue box is on the right of the robotic arm [0.052, 0.0, 0.552, 0.342].} While preserving the original language of the user's request, our system provides bounding boxes for identified objects in the scene. Additionally, with the robot positioned at the center of the image, we detail the relative positions of the referred objects, highlighting that the polar bear is below the robot and the blue box is to its right. The overview can be found at the top of Figure~\ref{fig:oci_train}.

\noindent
\textbf{Architecture.} We have selected the pre-trained ViT-L/14~\cite{vit} from CLIP~\cite{clip} as our visual encoder and adopted LLaMA-2-7B~\cite{llama, llama-2} as our LLM. To ensure modal alignment and provide the appropriate input dimension for the LLM, a fully connected layer is implemented to transform the ViT's $16 \times 16$ output embedding $V \in \mathbb R^{16 \times 16 \times 1024}$ to $V^{'} \in \mathbb R^{256 \times 4096}$. We tap into the robust vision-language capabilities inherent to the text-image alignment~\cite{minigpt-4}. Additionally, we fine-tune the associated networks while keeping the language and visual embeddings static, with only the alignment layers being adjustable. 

\begin{figure}[t]
    \centering
    \includegraphics[width=0.98\columnwidth]{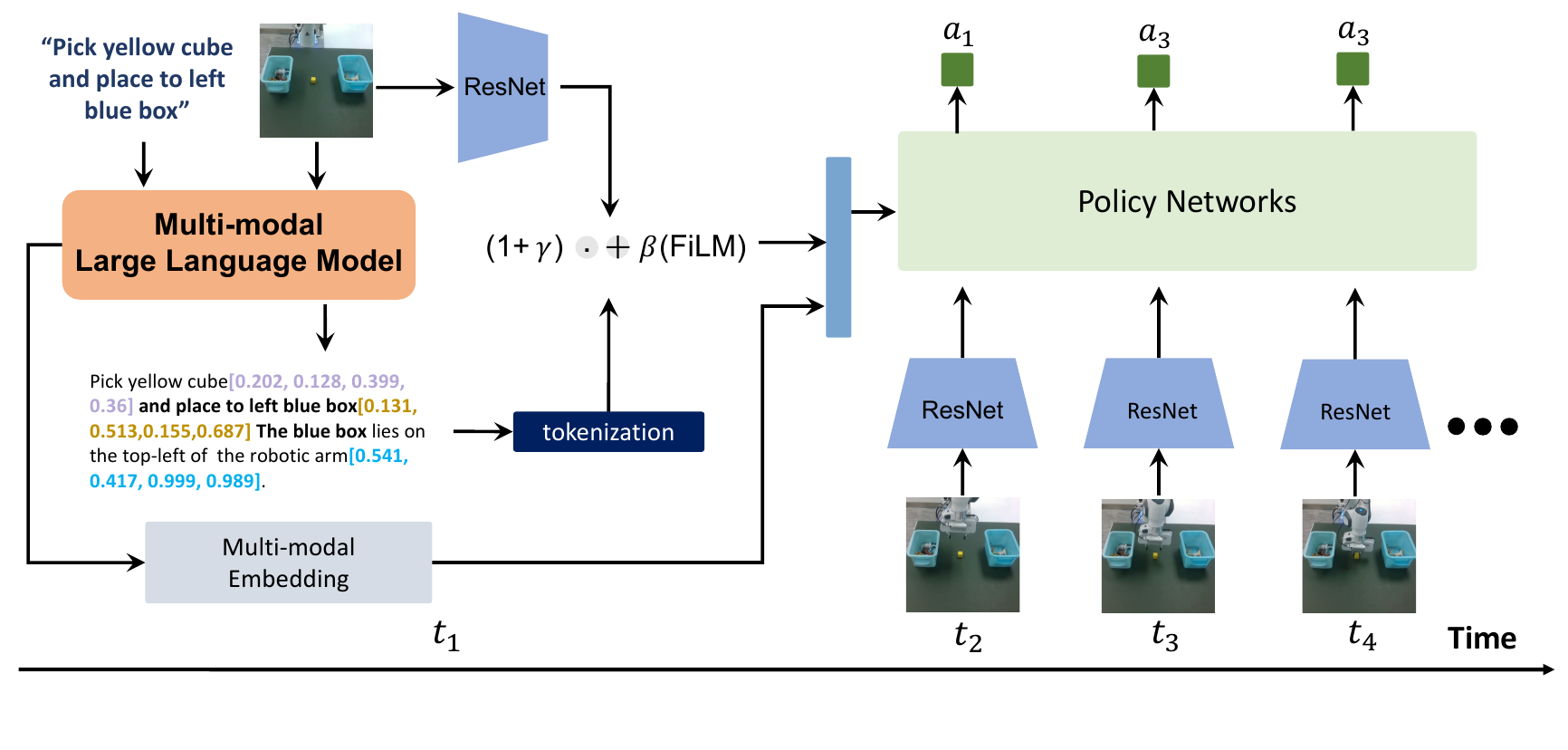}
    \caption{This figure illustrates how we connect the policy network with MLLM. During the fine-tuning phase, MLLM is only utilized at the initial time step $t_{1}$, and its parameters are frozen.}\label{fig:inference}
\end{figure}

\subsection{Feature Reuse Mechanism}
This section gives a detailed description of how we reuse the features from MLLM to improve the generalization of manipulation policy learning.

\noindent
\textbf{Policy networks.} To formulate an efficient multi-task robot policy, we employ policy networks designed with a multi-task decoder architecture. Concretely, our objective is to learn a robot policy represented by $\pi(a_{t}|P, H)$, where $H := \{o_{1}, a_{1}, o_{2}, a_{2}, \cdots, o_{t}\}$ captures the historical trajectory of past interactions. Within this framework, the $o_{t} \in \mathbb O$ and $a_{t} \in \mathbb A$ respectively represent observations and actions taken at each interaction step. These policy networks ingest multi-modal tokens. For encoding, we deploy multi-modal prompts: the image undergoes processing via a vision backbone and is subsequently combined with the tokenized, augmented instruction using straightforward concatenation. The next step for the policy networks is to delineate the action space. Our policy network consists of three MLP layers with ReLU non-linear activation.

\begin{figure*}[t]
    \centering
    \includegraphics[width=\textwidth]{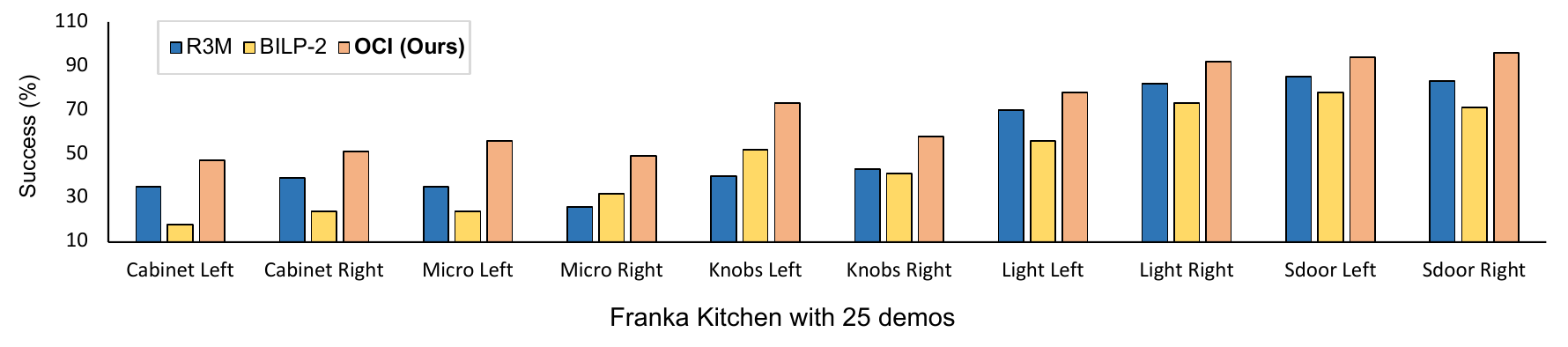}
    \caption{The experimental results on Franka Kitchen. On all sub-tasks, our proposed OCI beats existing approaches, where our methods lead for a large margin on some tasks.}
    \label{fig:franka_exp}
\end{figure*}
\begin{figure}[t]
    \centering
    \includegraphics[width=0.45\textwidth]{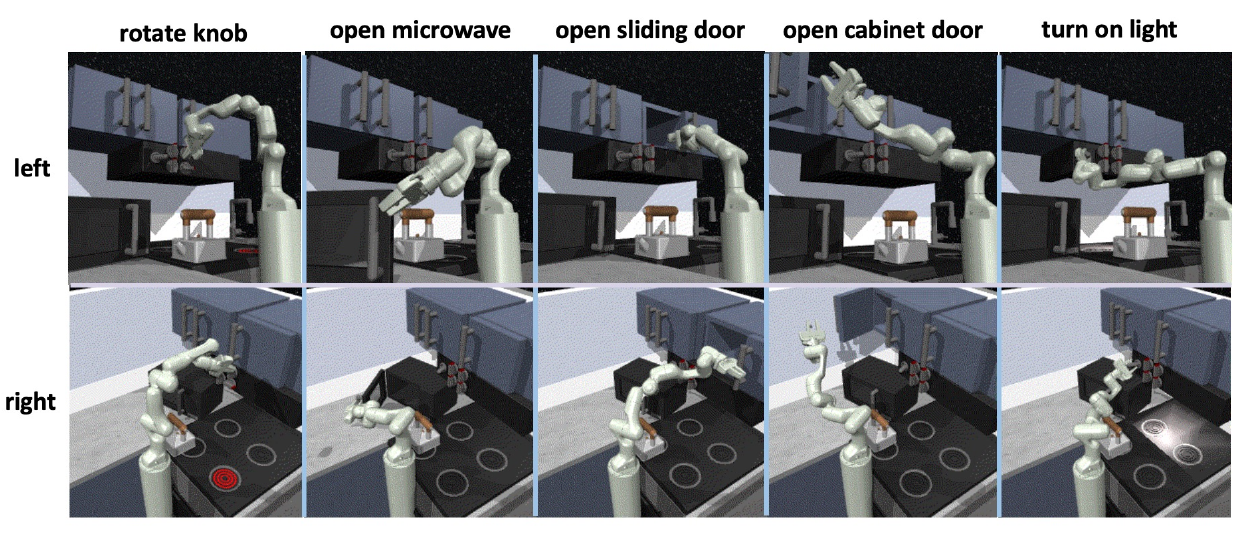}
    \caption{The example of Franka Kitchen for five tasks on two camera views.}
    \label{fig:frank_exp_setup}
\end{figure}

\noindent
\textbf{Feature Reuse Method.} In our setup, we primarily retrieve the augmented instruction at the first frame and then abandon the model. Yet, discarding the computations made within the MLLM once a new instruction is acquired is inefficient. This is because the MLLM provides not just vision-language comprehension of the original instruction but also intuitively recognizes an object's identity and position. Finding a method to harness this wealth of information is both challenging and rewarding. 

To implement this idea, we architected a feature reuse mechanism. In specific, we denote the feature embedding by the final layers of the MLLM as $E_{mllm} \in \mathbb R^{D}$ and define the multi-modal token (fusion of image tokens and text tokens) as $M \in \mathbb R^{D'}$, where $D'$ and $D$ are its corresponding dimensions. Herein, we add a sequence of operations with a LayerNorm-MLP (LN-MLP)~\cite{ba2016layer} ($\mathbb R^{D} \rightarrow \mathbb R^{D'}$), such that $E_{mllm}^{'} = \text{LN-MLP}(E_{mllm} \in \mathbb R^{D'})$. The LN-MLP introduces nonlinearity and allows more flexible transformations for the $E_{mllm}$. Here, the $E_{mllm}^{'}$ carried prior knowledge on vision-language understanding. Therefore, it is paramount to align this prior vision-language information with the current instruction-observation pair. We use cross-attention to serve as a bridge to connect these two types of multi-modal embedding.


Particularly, $M$ is projected into a query (Q) and key (K), and $E_{mllm}^{'}$ is projected into value (V). The keys $K$ and values $V$ are down-sampled to different sizes for different heads indexed by $i$. Thus, we formulate our multi-scale cross-attention (MSC) as $Q_{i} = E_{mllm}^{'}W_{i}^{Q}$, $K_{i} = MSC(M, r_{i})W_{i}^{K}, V_{i} = MSC(M, r_{i})W_{i}^{V}$, and $V_{i} = V_{i} + P(V_{i})$. The $MSC(\cdot, r_{i})$ is an MLP layer for aggregation in the $i^{\text{th}}$ head with the down-sampling rate of $r_{i}$, and $P(\cdot)$ is a depth-wise convolutional layer for projection. Compared with the standard cross-attention, more fine-grained and low-level details that are beneficial to manipulation tasks are preserved. Finally, we calculated the attention tensor by:
\begin{equation}
    h_{i} = Softmax(\frac{Q_{i}K^{T}_{i}}{\sqrt{d_{h}}}V_{i})
\end{equation}
where $d_{h}$ is the dimension. Intuitively, the instruction-observation queries the useful knowledge in vision-language knowledge embedding and passes it into the policy network to improve the successful rate on unseen domains. 

Note that the MLLM is frozen during the policy learning. At test time, we extract both the augmented instruction and feature embedding at the task's initial frame. These instructions and feature embedding are then stored for subsequent frame inferences, allowing the MLLM to be purged from the processor's memory. This optimizes computational speed by freeing up resources.

\noindent
\textbf{Overview of the Framework.} In our context, the text includes not only natural language but also regions of interest that are represented by a set of floating points. Inputting regions of interest into the model includes various approaches, such as direct concatenation with cropped image patches~\cite{bracha2023disclip, zhou2023make}, encoder-decoder structure for bounding boxes representation~\cite{lg3d, lgd, zhu2022teach, zhu2023scalekd}, and utilization of Gaussia map~\cite{lin2020interactive, lin2022multi, hao2020labelenc}. In our experiment, we find it is sufficient to directly incorporate bounding boxes as a natural language without additional pre-processing. Initially, a pre-trained image encoder processes each image. This is then tokenized, similar to RT-1~\cite{brohan2022rt1}. The language undergoes tokenization first, after which a T5-small model extracts features. It is then concatenated with the image token. A policy network is followed up to generate the action space. A comprehensive overview of this framework can be found in Figure~\ref{fig:inference}.

\section{Experiments}
Our experiments aim to answer the following questions: 1) Does our method enable better policy learning than using naive language instruction? 2) Is our method effective in real-world environments? We initiate our discussion by outlining the simulation environments tailored to address these queries. Subsequently, we present in-depth experimental results that positively affirm answers to both questions.
\begin{table}[t]
\caption{Ablation on two types of augmented instruction. Experiments are conducted on Franka Kitchen. \enquote{w/o} indicates \enquote{without}. We report the average success rate over five tasks and two camera views per task.}
\centering
\begin{tabular}{lcc}
\toprule
Model & 10 Demos & 25 Demos\\
\midrule
OCI (Ours) &  61.7 &  69.4\\
 w/o absolute position & 54.2 & 63.5 \\
 w/o relative position & 49.6 & 57.3\\
\bottomrule
\end{tabular}
\label{table:position_ablation}
\end{table}

\begin{table}[t]
\caption{Ablation on reusable feature mechanism. Experiments are conducted on Franka Kitchen. \enquote{w/o} indicates \enquote{without}. We report the average success rate over five tasks and two camera views per task.}
\centering
\begin{tabular}{lcc}
\toprule
Model & 10 Demos & 25 Demos\\
\midrule
OCI (Ours) & 61.7 &  69.4\\
w/o reusable features & 53.9 & 62.3\\
\bottomrule
\end{tabular}
\label{table:reusable_ablation}
\end{table}

\subsection{Simulation Experiments}
\noindent
\textbf{Experiments Setup.} Franka Kitchen benchmark focuses on tasks like sliding open the right door, opening the cabinet, turning on the light, turning the stovetop knob, and opening the microwave. The example of each task is present in Figure~\ref{fig:frank_exp_setup}. For Franka Kitchen tasks, the length of a demonstration is 50, which contains 50 state-action pairs. For MLLM, we fixed the weights when training the policy networks. There are two views in the Franka Kitchen simulator; each is conducted three times, and the average success rate over both views is reported. All Franka tasks include proprioceptive data of the arm joint and gripper positions. The horizon for all Franka tasks is 50 steps, and our imitation experiments use either 10 or 25 demos.

\noindent
\textbf{Baselines}. We compare our model with R3M~\cite{nair2022r3m}, which is the state-of-the-art method and widely applicable method in Franka Kitchen. We also compare with BLIP-2~\cite{bracha2023disclip}, a SOTA vision-language model. We replace our MLLM with BLIP-2 in our method and retain the FRM. In all experiments, the policy network is learned using few-shot learning on a small amount of demonstration data. There are two settings, one of which utilizes 25 demonstrations, and the other utilizes ten demonstrations. We report the success rate in five tasks per benchmark and two different camera views for each set respectively. Each experiment is run five times, and we report the average performance. 

\noindent
\textbf{Main Experimental Results.} We demonstrate the experimental results in Figure~\ref{fig:franka_exp}. It is obvious that for all tasks on two different camera views, our approach achieves superior performance over both R3M and BLIP-2. On some difficult tasks, such as opening the cabinet, opening the microwave, and turning the stovetop knob, the performance gap between our proposed OCI is even larger compared to R3M and BILP-2. Compared to BILP-2, which also uses a large-scale pre-trained vision-language model to align the instruction and observation, then map to the policy network, our OCI achieves stronger. performance over five tasks, showing the effectiveness of the object-centric instruction augmentation. 
\begin{figure*}[t]
    \centering
    \includegraphics[width=\textwidth]{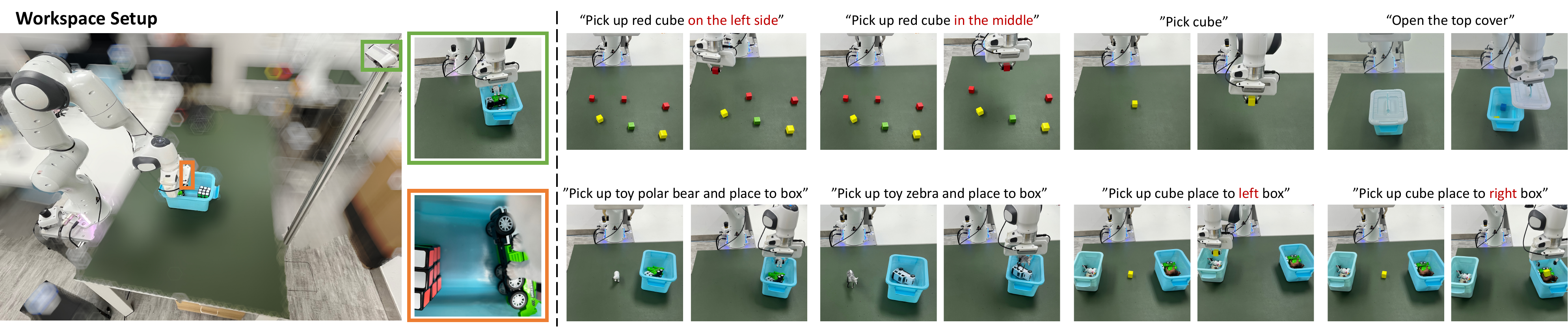}
    \caption{\textbf{Left:} The setup of our Franka real robot. \textbf{Right:} The example of some tasks that we collected.}\label{fig:exp_setup}
\end{figure*}

\begin{figure}[t]
    \centering
    \includegraphics[width=\columnwidth]{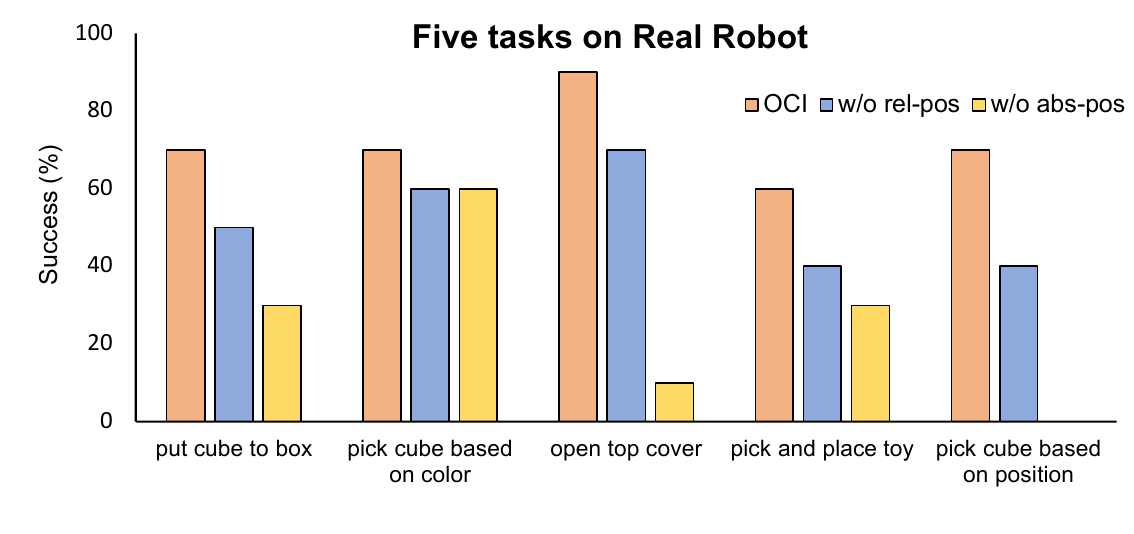}
    \caption{Results from real-world experiments indicate that both relative and absolute positions are crucial for the successful execution of manipulation tasks.}\label{fig:real_robot_exp}
\end{figure}
\noindent
\subsection{Ablation Study}

\noindent
\textbf{How Important for Absolute and Relative Position in Instruction?} 
We evaluate the utility of instruction augmentation and present our findings in Table~\ref{table:position_ablation}. Our experiments indicate that both relative and absolute positions are critical, confirming the efficacy of our framework. We conduct another ablation study is conducted on real-world experiments. 

\noindent
\textbf{The Effectiveness of Feature Reuse Mechanism.}
We explore the efficacy of our proposed FRM. The experimental results are detailed in Table~\ref{table:reusable_ablation}. Observing the data, we find that omitting the FRM results in an average drop of 7.1\% in the success rate, highlighting the importance of FRM.

\subsection{Real-world Experiments} 
Given our encouraging simulation experiments, it is natural to ask whether our algorithm can work on real-world robotic manipulation tasks, which present the additional challenges of noisy image observations from imperfect camera sensors and increased object quantity and diversity. 

We use the Franka robot with a 7-DOF arm, which is equipped with a parallel jaw gripper (see Figure~\ref{fig:exp_setup}, left). Our workspace uses two D435i RealSense RGBD cameras. We only use the RGB information in our experiments. One egocentric camera is attached to the robot's hand, and one exocentric camera is positioned at the robot's front. 

\noindent
\textbf{Experiment Setup.} To realize the stated challenges above, we design a real-world environment (referred to as RealRobot), in which a Franka robot is tasked with 1) picking up a cube to either left or right box, 2) picking up a cube based on the color, 3) open the top cover of a box, 4) pick up a toy and place to the box, 5) pick up a cube based on its position (left, middle, or right). We provide a sample of the data we gathered from real-world trajectories in Figure~\ref{fig:exp_setup} (right). The tasks are numerically represented for simplicity; for example, Task 1 corresponds to \enquote{picking up a cube and placing it in either the left or right box}.

\noindent
\textbf{Experimental Results.} An evaluation of our approach was conducted using a real robot setup, with each task repeated in ten trials. Additionally, we conducted an ablation study, which is presented in Figure~\ref{fig:real_robot_exp}. In this study, we progressively removed the relative position (denoted as \enquote{w/o rel-pos}) and the absolute position (denoted as \enquote{w/o abs-pos}). Across all five tasks, our OCI consistently outperformed the baselines, achieving the highest success rates. Particularly notable were the results for Task 1 and Task 5. In these tasks, the robot needed to discern direction, such as picking up the cube from the left, right, or middle side or placing the cube on the left or right box. Plain language instructions struggled with Task 5, failing entirely, and achieved a mere 30\% success rate for Task 1. However, introducing absolute position information boosted success rates by 40\% and 20\% for the two tasks, respectively. Incorporating relative position data further enhanced performance by 20\% and 30\%, underscoring the effectiveness of our OCI.

\section{Conclusion}
This work contributes a novel perspective on language instruction for robotics manipulation. Motivated by the concept of visual understanding in human intelligence, we augment language instruction by adding an object's absolute and relative position into the text format. Such augmented language alleviates the burden of the visual encoder, which was previously responsible for localizing objects on its own. We conduct experiments on both simulation and real-world scenarios and show the superior performance of our methods over conventional language instruction. We believe our approach presents a fresh perspective on the kinds of instructions best suited for versatile robotic manipulation.

\clearpage
\bibliographystyle{IEEEtran}
\bibliography{reference}

\end{document}